\documentclass[
	runningheads
]{article}
\usepackage{arxiv}

\usepackage[utf8]{inputenc} 
\usepackage[T1]{fontenc}    

\usepackage{listings}
\usepackage{graphicx} 
\usepackage{emile}

\usepackage{amsmath}
\usepackage{amssymb}
\usepackage{mathtools}
\usepackage{amsthm}
\usepackage[american]{babel}
\usepackage{hyperref}
\usepackage{stmaryrd}
\usepackage[normalem]{ulem}
\usepackage{algorithm}
\usepackage{algpseudocode}

\newcommand{\method}{Unified Language for LEarning and Reasoning}
\newcommand{\langname}{\textsc{ULLER}}

\newcommand{\nso}{\llbracket}
\newcommand{\nsc}{\rrbracket}
\newcommand{\nesysystem}{{\nso \nsc}}

\newcommand{\nesyoutput}{\mathcal{B}}
\newcommand{\uniondomain}{\mathcal{O}}

\newcommand{\gramsep}{\ | \ }
\newcommand{\variables}{\mathcal{V}}
\newcommand{\predicates}{\mathcal{P}}
\newcommand{\properties}{\mathcal{T}}
\newcommand{\functions}{\mathcal{F}}

\newcommand{\constants}{\mathcal{C}}
\newcommand{\domains}{\mathcal{D}}

\newcommand{\nesylanguage}{\mathcal{L}_{\text{\langname}}}

\newcommand{\classic}{\nsc^ \text{C}}
\newcommand{\fuzzy}{\nsc^ \text{\normalfont F}}
\newcommand{\nesyfuzzy}{\nsc^ \text{\normalfont NeSy}}
\newcommand{\probabilistic}{\nsc^ \text{\normalfont P}}
\newcommand{\nscsampling}{\nsc^\text{\normalfont S}}
\DeclareMathOperator{\tnorm}{\otimes}
\DeclareMathOperator{\tconorm}{\oplus}
\newcommand{\assign}{{\mathcal{A}}}
\newcommand{\prob}{{p}}
\newcommand{\eval}{\mathrm{true}}
\newcommand{\limplies}{\Rightarrow}
\newcommand{\lequiv}{\Leftrightarrow}

\newcommand{\domainspace}{\Omega}

\newcommand{\interp}{I}
\newcommand{\interpretation}{\interp}

\newcommand{\atom}{a}

\newcommand{\finex}{\hspace*{\fill}$\triangleleft$}

\newcommand{\RM}[1]{\textcolor{magenta}{\textbf{[RM: #1]}}}
\newcommand{\EK}[1]{\textcolor{brown}{\textbf{[EvK: #1]}}}
\newcommand{\SB}[1]{\textcolor{blue}{\textbf{[SB: #1]}}}
\newcommand{\EG}[1]{\textcolor{orange}{\textbf{[EG: #1]}}}

\renewcommand{\RM}[1]{}
\renewcommand{\EK}[1]{}
\renewcommand{\SB}[1]{}
\renewcommand{\EG}[1]{}

\theoremstyle{plain}
\newtheorem{theorem}{Theorem}[section]

\theoremstyle{definition}
\newtheorem{definition}[theorem]{Definition}

\theoremstyle{remark}

\theoremstyle{definition}
\newtheorem{example}[theorem]{Example}

\title{\langname{}: A Unified Language\\ for Learning and Reasoning}
\renewcommand{\shorttitle}{\langname{}: A Unified Language for Learning and Reasoning}

\author{ \href{https://orcid.org/0000-0001-5502-4817}{\includegraphics[scale=0.06]{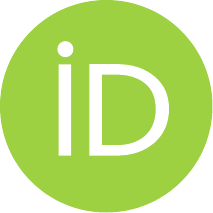}\hspace{1mm}Emile van~Krieken}$^\ast$ \\
	University of Edinburgh \\
	\And
	\href{https://orcid.org/0000-0003-1624-9188}{\includegraphics[scale=0.06]{figs/orcid.pdf}\hspace{1mm}Samy Badreddine}$^\ast$\\
	Sony AI \\
        Fondazione Bruno Kessler \\
        UniTrento \\
        \And
        \href{https://orcid.org/0000-0001-9907-7486}{\includegraphics[scale=0.06]{figs/orcid.pdf}\hspace{1mm}Robin Manhaeve} \\
	KU Leuven\\
        \And
        \href{https://orcid.org/0000-0001-9313-753X}{\includegraphics[scale=0.06]{figs/orcid.pdf}\hspace{1mm}Eleonora Giunchiglia} \\
	TU Wien
}
\date{May 1, 2024}

\begin{document}

\maketitle
\begin{abstract}

    The field of neuro-symbolic artificial intelligence (NeSy), which combines learning and reasoning, has recently experienced significant growth.
    There now are a wide variety of NeSy frameworks, each with its own specific language for expressing background knowledge and how to relate it to neural networks. 
    This heterogeneity hinders accessibility for newcomers and makes comparing different NeSy frameworks challenging.
    We propose a unified language for NeSy, which we call \langname{}, a \method{}.
    \langname{} encompasses a wide variety of settings, while ensuring that knowledge described in it can be used in existing NeSy systems.
    \langname{} has a neuro-symbolic first-order syntax for which we provide example semantics including classical, fuzzy, and probabilistic logics. 
    We believe \langname{} is a first step towards making NeSy research more accessible and comparable, paving the way for libraries that streamline training and evaluation across a multitude of semantics, knowledge bases, and NeSy systems.
\end{abstract}
\def\thefootnote{*}\footnotetext{Equal contribution. \\ Correspondence to \url{Emile.van.Krieken@ed.ac.uk} and \url{samy.badreddine@sony.com}
}\def\thefootnote{\arabic{footnote}}

\section{Introduction}

Deep learning has driven innovation in many fields for the past decade. Among the many reasons behind its central role is the ease with which it can be applied to a multitude of problems. Recently, neuro-symbolic (NeSy) methods (see, e.g.,~\cite{manhaeve2018deepproblog,badreddineLogicTensorNetworks2022,NEURIPS2023_4d9944ab,GIUNCHIGLIA2024109124,pryorNeuPSLNeuralProbabilistic2023,yangNeurASPEmbracingNeural2020,huangScallopProbabilisticDeductive2021}), 
which belong to the NeSy subfield \emph{informed machine learning},~\cite{giunchigliaDeepLearningLogical2022,vonruedenInformedMachineLearning2023} 
have overcome some well-known problems affecting deep learning models by exploiting \emph{background knowledge} available for the problem at hand. For example, knowledge can help in training models with fewer data points and/or incomplete supervisions, creating models that are compliant by-design with a set of requirements, and being more robust in out-of-distribution prediction. 

However, the presence of background knowledge makes it more challenging to obtain ``frictionless reproducibility''~\cite{donoho2023data} which characterises machine learning (ML). Indeed, in ML, shared datasets and clear evaluation metrics allow ML practitioners to quickly get started with evaluating new methods and comparing it to existing work. 
To achieve this goal for NeSy research, we also need \emph{frictionless sharing of knowledge}. 
Current NeSy frameworks all have different approaches to encode the background knowledge: some use logical languages, like first-order \cite{badreddineLogicTensorNetworks2022,pryorNeuPSLNeuralProbabilistic2023}, propositional \cite{ahmed2022spl,xu2018semanticloss,GIUNCHIGLIA2024109124}, logic programming \cite{manhaeve2018deepproblog} or answer set programming \cite{yangNeurASPEmbracingNeural2020} - with a wide array of different syntaxes - while other methods use plain Python programs \cite{de2024differentiable}. See Section~\ref{sec:related-work} for an overview. 
To compare the performance of different NeSy systems, a researcher needs to specify the same knowledge in many languages. This is a significant barrier for researchers new to the field, and, even for experts, is a time-consuming and error-prone task. 

\subsubsection{\langname{}, a \method{}}
We take a first step towards frictionless sharing of knowledge in the NeSy field by proposing a \emph{Unified Language for Learning and Reasoning} (\langname{}, pronounced ``OOH-ler'' like the god of the Norse mythology). 
\langname{} allows us to express the knowledge used in informed machine learning. The long-term goal is to create a Python library implementing \langname{} to be shared among the significant NeSy systems. 
First, the user expresses the knowledge in \langname{}. Then, they load the data, after which they call different NeSy systems with a single line of code to train neural networks, or to use the knowledge at prediction time. This allows the NeSy community $(\mathit{i})$ to define benchmarks that include both data and
knowledge, $(\mathit{ii})$ to easily compare the available NeSy systems on such benchmarks, and $(\mathit{iii})$ to lower the barrier to entry into NeSy research for the broader machine learning community. 

To achieve the above requires decoupling the \emph{syntax} of the knowledge representation from the \emph{semantics} given by the NeSy system. 
The syntax of \langname{}, defined in Section~\ref{sec:syntax}, is syntactic sugar around first-order logic (FOL), with specifically designed \emph{statement bindings}. Statements simplify the process of writing down function application and composition - and hence dealing with data sampling and processing pipelines. 
We opt for a FOL-like syntax. It generalises propositional logic, while being a common language for declaring general constraints. We adapt the FOL syntax to one that is more familiar to ML researchers, who are mostly used to writing procedural statements like in Python, while having a well-defined semantics for logicians. 
Furthermore, FOL is highly expressive: We believe that it can express all knowledge currently used in NeSy methods. 

The semantics of \langname{} (Section~\ref{sec:semantics}), depends on $(\mathit{i})$ an \emph{interpretation}, often referred to as a ``symbol grounding'' \cite{harnadSymbolGroundingProblem1990}, which maps symbols to meanings, and $(\mathit{ii})$ a \say{NeSy system}, which takes knowledge and its interpretation, and computes loss functions and outputs accordingly. We formalise the differences between NeSy systems by what they compute
given a program in \langname{} and an interpretation. We also provide examples for several common systems, such as classical logic, fuzzy logic (such as Logic Tensor Networks~\cite{badreddineLogicTensorNetworks2022}), probabilistic logic (such as Semantic Loss~\cite{xu2018semanticloss} and DeepProbLog~\cite{manhaeve2018deepproblog}). This
highlights the flexibility of our language, as it can be used to express knowledge in many formalisms.

\section{Syntax of \langname{}}
\label{sec:syntax}
Let $\variables$ be a set of variable symbols, $\constants$ be a set of constant symbols, $\domains$ be a set of a set of domain symbols, $\predicates$ be a set of predicate symbols, $\properties$ be a set of property symbols, and $\functions$ be a set of function symbols. We then define the syntax of \langname{} $\nesylanguage$ as a context-free grammar:
\begin{equation}
\begin{aligned}
    \label{eq:syntax-nesy}
    F&::= \forall x \in D\ (F) \gramsep \exists x \in D\ (F)   \\
    F&::= F \land F \gramsep F \lor F \gramsep F\Rightarrow F \gramsep \neg F \gramsep \mathrm{P}(T, ..., T) \gramsep ( F ) \\
    F &::= x := f(T, ..., T)\ (F) \\
    T &::= x \gramsep c \gramsep T.\text{prop} \gramsep T + T \gramsep T-T \gramsep \dots  
\end{aligned}
\end{equation}
where $D \in \domains$, $x \in \variables$, $c \in \constants$, $f, +, - \in \functions$, $\mathrm{P} \in \predicates$ and $\text{prop}\in \properties$. The nonterminal symbol $F$ is a \emph{formula} and $T$ is a \emph{term}.
We call $x:= f(T,\dots,T)(F)$ a statement binding, or just \emph{statement}, which we discuss in Section~\ref{sec:statement}.
Notice that, except for basic arithmetic operations ($+$, $-$, \dots), functions only appear in statements.

The syntax of \langname{} does not include a special syntactic construct for neural networks. 
Instead, we treat them as functions, where the intended meaning is given by the semantics specified by the NeSy system. 
We therefore hide how the NeSy system uses the neural networks to the user, so the focus is on specifying constraints rather than implementation details.

\textbf{Syntactic Sugar.}  We use $\forall x_1 \in D_1, x_2 \in D_2 ( F )$ as syntactic sugar for $\forall x_1 \in D_2\ (\forall x_1 \in D_2\ (F))$ for the quantifiers. We also use $x_1 := f_1(T,\dots,T), x_2 := f_2(T, \dots, T)\ (F)$ as syntactic sugar for $x_1 := f_1(T,\dots,T)\ (x_2:= f_2(T, \dots, T)\ (F))$. Finally, we also allow for binary predicates in infix notation, such as $T \leq T$.

\textbf{Typing.} \langname{} is a dynamically typed language. We do not guarantee syntactically nor via type checker that functions and predicates only take arguments from the domain defined in their interpretations. This mimics the design of the type system of Python.

\subsection{Statements}
\label{sec:statement}
A key design choice of \langname{} is the use of special statement bindings $x := f(T, ..., T)\ (F)$ to declare (possibly random) variables obtained by applying (possibly non-deterministic) functions. 
The function symbols $f$ appear only in statements, and not in the definition of terms $T$, like in standard FOL. 
Statements simplify the composition of functions.
They give a syntax that is both familiar to ML researchers who are used to writing Python, and gives a clear separation between the machine learning pipeline that processes data and the constraints on the data given by the logic. 
We will motivate statements with the two following examples.

\begin{example}[Procedural composition of functions]
    \label{ex:composition}
    Consider the MNISTAdd example from Appendix~\ref{sec:examples}.
    To emphasise the ease of composing functions in \langname{}, consider a scenario where $(i)$ the two MNIST digits always represent different digits, and $(ii)$ the classifier $f$ expects greyscale images while the data points in the dataset $T$ are RGB images.
    We can easily apply transformations and formulate the new condition using \langname{} statements:
    \begin{align}
        \begin{split}
            & \forall x \in T (\\
            & \quad x_1' := \mathrm{greyscale}(x.\mathrm{im1}), x_1'' := \mathrm{normalise}(x_1'), \\
            & \quad x_2' := \mathrm{greyscale}(x.\mathrm{im2}), x_2'' := \mathrm{normalise}(x_2'), \\
            & \quad n_1 := f(x_1''), n_2 := f(x_2''), \\
            & \qquad ( (n_1 + n_2 = x.\mathrm{sum}) \land n_1 \neq n_2 ) \\
            & )
        \end{split}
    \end{align}
\finex
\end{example}

\def\dice{\mathrm{dice}}
\def\iseven{\mathrm{even}}
\def\isa{\mathrm{isa}}
\begin{example}[Scoping independence]
 Another key feature of \langname{} statements is that they explicitly delimit the scopes of variables, giving control over the memoisation and independence assumptions.
    Consider a non-deterministic function
    $\dice()$ which associates a probability to each outcome of a six-sided dice throw.
    Consider the following program written in \langname{}:
    \begin{equation}
        \label{eq:shared_throw}
        x := \dice()\ (x=6 \land \iseven(x)).
    \end{equation}
    The formula asks whether a die-throw outcome is both a six and even. 
    For a fair dice, the probability of the formula  is $\frac{1}{6}$ under probabilistic semantics.
    
    Now consider the alternative \langname{} program:
    \begin{equation}
        \label{eq:independent_throws}
        (x := \dice()\ (x=6)) \land (x := \dice()\ (\iseven(x))).
    \end{equation}
    In this program, we throw two independent dice, and check if the first lands on six and the second is even.
    For fair dice, the probability of this formula is $\frac{1}{6} \cdot \frac{1}{2}=\frac{1}{12}$.

    Consider a similar program in regular FOL (which is not allowed in \langname{}):
    \begin{equation}
        (\dice() = 6) \land \iseven(\dice())
    \end{equation}
    Here, it is ambiguous whether the outcomes of the dice are shared like in the \langname{} program of \eqref{eq:shared_throw} or not, like in \eqref{eq:independent_throws}.
    Many probabilistic NeSy frameworks choose the first option and memorise the outcome of the function.
    We argue that this behaviour should not be a default assumption from the NeSy system.
    Instead, dependence and memoisation scopes should be explicitly defined by the program.
    \langname{} statements give researchers control over these scopes.
\finex
\end{example}

\section{Semantics of \langname{}}
\label{sec:semantics}
In this Section, we define the semantics of \langname{}. In Section \ref{sec:interpretation} we discuss how \langname{} interprets the symbols in the language, such as the function and domain symbols. Then, in Section \ref{eq:neuro-symbolic-system}, we discuss how different NeSy systems interpret the formulas in \langname{}.
\subsection{Interpretation of the Symbols}
\label{sec:interpretation}

To assign meaning to \langname{} programs, we need to \emph{interpret} the non-logical symbols in $\nesylanguage$, that is, $\domains$, $\predicates$, $\functions$, and $\constants$, using an interpretation function $\interp$.

\begin{definition}
    \label{def:interpretation}
    An interpretation $\interp$ is a function assigning a meaning to the symbols in $\nesylanguage$ under the following rules, where $\domainspace_1, ..., \domainspace_{n}, \domainspace_{n+1}$ are sets. 
\begin{enumerate}
    \item The interpretation of a domain $D\in \domains$ is a set $\domainspace$.
    \item The interpretation of a predicate $\mathrm{P}$ of arity $n$ is a function of $n$ domains to $\{0, 1\}$. That is,  $\interp(\mathrm{P}): \domainspace_1 \times ...\times \domainspace_n \rightarrow \{0, 1\}$. 
    \item The interpretation of the predicate $\eval\in \mathrm{P}$ is the identity function on $\{0, 1\}$, that is, $\interp(\eval): \{0, 1\} \rightarrow \{0, 1\}$ such that $\interp(\eval)(x)=x$.  
    \item The interpretation of a constant $c$ is an element of a domain $I(c) \in \domainspace_i$.
    \item The interpretation of a function $f$ of arity $n$ is a \emph{conditional probability distribution}\footnote{To be precise, our definition is equivalent to a \emph{probability kernel} or \emph{Markov kernel}, which is a formalisation of the concept of a conditional probability distributions in measure theory.} $I(f): \domainspace_1 \times  ... \times \domainspace_n \rightarrow (\domainspace_{n+1} \rightarrow [0, 1])$. That is, for any set of inputs $x_1\in \domainspace_1, ..., x_n\in \domainspace_n$, $I(f)(x_1, ..., x_n)$ is a probability distribution on the domain $\domainspace_{n+1}$. If for all $x_1 \in \domainspace_1, ..., x_n \in \domainspace_n$ the probability distribution $I(f)(x_1, ..., x_n)$ is a deterministic distribution, we say that $I(f)$ is a \emph{deterministic function}. 
\end{enumerate}
\end{definition}

We give a probabilistic interpretation to both domains and functions. 
In particular, we treat functions, such as neural networks, as a \emph{conditional distribution} given assignments $x_1\in \domainspace_1, ..., x_n\in\domainspace_n$ to input variables. 
This allows us to represent the uncertainty of the neural networks, which NeSy systems using, for example, probabilistic and fuzzy semantics can use to compute probabilities and fuzzy truth values. 
We will also frequently want to use regular (deterministic) functions $f: \domainspace_1 \times ... \times \domainspace_n \rightarrow \domainspace_{n+1}$. A regular function is a special case of a conditional distribution that we refer to as a deterministic function. We define deterministic functions with a conditional distribution using the Dirac delta distribution at $f(x_1, ..., x_n)$ for continuous distributions, and a distribution that assigns 1 to the output value $f(x_1, ..., x_n)$ for finite domains, and 0 to the other values.

\subsection{Semantics of neuro-symbolic systems}
\label{eq:neuro-symbolic-system}
\EK{Renamed this section from 'interpretation of formulas}

\begin{figure}
    \centering
    \includegraphics[width=.8\textwidth]{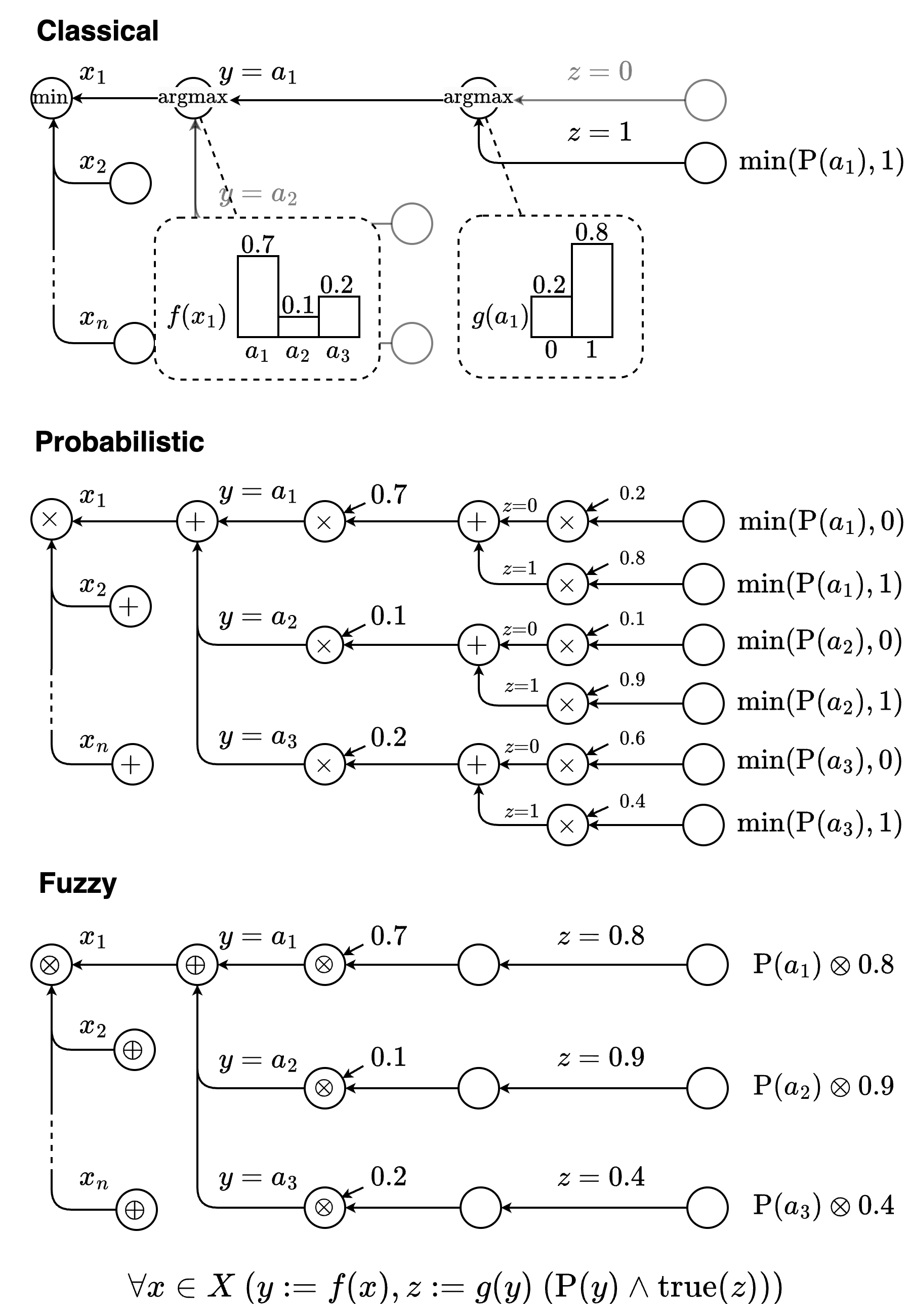}
    \caption{The meaning of an example \langname{} formula under classical, probabilistic and fuzzy semantics. We interpret the function symbols as conditional distributions $f: \{x_1, ..., x_n\} \rightarrow (\{a_1, a_2, a_3\} \rightarrow [0,1])$ and $g: \{(a_1, a_3, a_3\} \rightarrow (\{0, 1\} \rightarrow [0,1])$. With abuse of notation, we ignore $I()$ and $\nso \nsc$.}
    \label{fig:semantics_flow}
\end{figure}

We next define the meaning of a formula in $\nesylanguage$, which requires both an interpretation $\interpretation$ and a \emph{NeSy system} $\nesysystem$. Here, $\nesysystem$ is a function that interprets the semantics of the program statements in $\nesylanguage$. We also need a variable assignment $\eta: \variables \rightarrow \uniondomain$ that maps variables $v\in \variables$ to an element of a domain $\uniondomain=\cup_i \domainspace_i$, where $\domainspace_i=I(D_i)$ is a set \RM{only time we use the term sample space, use set instead?}associated to a domain $D_i \in \domains$.
\EK{This is somewhat imprecise because variables can be the output of a function! Then it is not part of a domain set.}

\begin{definition}
    \label{def:nesystruct}
	A \emph{NeSy structure} is a tuple $(\interpretation, \eta, \nesyoutput, \nesysystem_{\interpretation, \eta})$ where $\interpretation$ is an interpretation, $\eta: \variables\rightarrow \uniondomain$ is a variable assignment, $\nesyoutput$ is a set of outputs and $\nesysystem_{\interpretation, \eta}: \nesylanguage \rightarrow \nesyoutput\cup \uniondomain$ is a \emph{neuro-symbolic system} which is a function that assigns an output in $\nesyoutput$ to each formula in $\nesylanguage$ and a domain element in $\uniondomain$ for terms $T$. If the interpretation and variable assignment are clear from the context, we write $\nesysystem$ for $\nesysystem_{\interpretation, \eta}$.
\end{definition}

We discuss several NeSy systems and their semantics for the NeSy language in the following sections, and provide a visual overview in Figure~\ref{fig:semantics_flow}. Each NeSy system is defined over some set of outputs $\nesyoutput$. For example, classical logic is defined over the output $\{0, 1\}$, while fuzzy logics are defined over the interval $[0, 1]$. A neuro-symbolic system $\nesysystem_{\interpretation, \eta}$ defines the semantics of a language expression. When a language expression is a term $T$, $\nesysystem_{\interpretation, \eta}$ returns an element of the universe $\uniondomain$. When the language expression is a formula $F$, it returns an element in $\nesyoutput$. 



\textbf{Notation.} We use $\eta[x \mapsto a]$ to update a variable assignment $\eta$ with the assignment of $a$ to $x$:
\begin{equation}
    \begin{aligned}
        \eta[x \mapsto a](x) &= a\\
        \eta[x \mapsto a](x') &= \eta(x') \quad \text{for}\ x' \neq x
    \end{aligned}
\end{equation}


We also define $\prob_f(a|T_1, ..., T_n)=\interpretation(f)(\nso T_1\nsc, ..., \nso T_n\nsc)(a)$, which computes the probability of the element $a\in \domainspace_{n+1}$ under the distribution $I(f)$, conditioned on the interpretation of the terms $T_1$ to $T_n$. That is, under $\nso T_1\nsc, ..., \nso T_n\nsc$. In the coming sections, we will frequently use this shorthand to talk about the semantics of the different NeSy systems. 

\subsection{Classical semantics}
We first define the semantics of the NeSy language if we \say{choose} an option deterministically from a conditional distribution. Then, under the classical semantics of the logical symbols, \langname{} is a regular first-order logic. 
A common way to make a deterministic choice from a distribution is to take its mode, that is, the most likely output according to the neural network. 

\begin{definition}
    \label{def:classical-semantics}

    The classical structure $(\interpretation, \eta, \{0, 1\}, \nso\classic_{\interpretation, \eta})$ is defined on boolean outputs $\{0, 1\}$ as:
    \begin{align}
            \label{eq:classic-universal}
            \nso \forall x \in D\ (F) \classic_{\interpretation, \eta} &= \min_{a \in I(D)}\nso F\classic_{\interpretation, \eta[x \mapsto a]} \\
            \nso \exists x \in D\ (F) \classic &= \nso \neg \forall x \in D\ (\neg F) \classic \\
            \nso F_1 \wedge F_2\classic = \min(\nso F_1\classic, \nso F_2\classic)&,\ \nso F_1 \lor F_2 \classic = \nso \neg (\neg F_1 \land \neg F_2)\classic \\
            \label{eq:negation}
            \nso \neg F\classic = 1 - \nso F\classic&,\          
            \nso F_1 \Rightarrow F_2 \classic = \nso \neg F_1 \lor F_2 \classic  \\         
            \nso P(T_1, \dots, T_n)\classic_{\interpretation, \eta} &= \interpretation(P)(\nso T_1\classic, \dots, \nso T_n\classic) \\
            \label{eq:constants}
            \nso x\classic_{\interpretation, \eta} = \eta(x)&,\ \nso c\classic = \interpretation(c) \\
            \nso T_1 + T_2\classic &= \nso T_1\classic + \nso T_2\classic\\
            \label{eq:property}
            \nso T.\mathrm{prop}\classic &= \mathrm{get}(\nso T\classic, \mathrm{prop}) \\
            \label{eq:classic-statement}
            \nso x := f(T_1, ..., T_n)(F) \classic_{I, \eta} &= \nso F\classic_{\interpretation, \eta[x \mapsto \argmax_{a\in \domainspace_{n+1}}\prob_f(a|T_1, ..., T_n)]} 
    \end{align}
    In Equation~\ref{eq:property}, $\mathrm{get}(\nso T\classic, \mathrm{prop})$ is a deterministic function that retrieves the value of an object property.
\end{definition}
Equation~\ref{eq:classic-statement} demands some explanation. The $\arg\max$ takes the probability distribution given by the interpretation of the function $f$ and chooses a value from the codomain $\domainspace_{n+1}$. In the classical structure, this choice is made deterministically by picking the mode of the distribution: the most likely element $a$. Then we assign this element $a$ to the variable $x$ through the variable assignment $\eta[x\mapsto a]$, and evaluate the rest of the formula $F$ under this new assignment.

Importantly, the classic semantics allows us to prove whether a neuro-symbolic system \say{is faithful} to classical logic when all functions are deterministic. We formally introduce this notion by noting we can transform any program into a deterministic program by choosing the mode of the distribution like in Equation~\ref{eq:classic-statement}.

\begin{definition}
    \label{def:classical-in-limit}
    For some interpretation $I$, the \emph{mode interpretation} $\hat{\interpretation}$, is another interpretation such that for all function symbols $f\in \functions$, $\hat{\interpretation}(f)$ returns the mode of $\prob_f$. That is, $\hat{p}_{f}(a|T_1, ..., T_n)= \delta(a-\arg\max_{a'} \prob_f(a'|T_1, ..., T_n))$, where $\delta$ is the Dirac delta distribution.
    Then a neuro-symbolic system $\nesysystem$ is \emph{classical in the limit} if for all language statements $L\in \nesylanguage$, $\nso L \nsc_{\hat{\interpretation}, \eta} = \nso L \classic_{\interpretation, \eta}$.
\end{definition}

\subsection{Probabilistic Semantics}
\label{sec:probabilistic-semantics}
Probabilistic semantics, also known as weighted model counting or possible world semantics in the literature, computes the probability that a formula is true. This is done by iterating over all possible assignments to the variables.

In the next sections, we will not redefine semantics whenever it is equal to the classical semantics, up to domain differences. For instance, we will not repeat constants and variable semantics.
\begin{definition}
    \label{def:probabilistic-semantics}
    The probabilistic structure $(\interpretation, \eta, [0,1],  \nso\probabilistic)$ is defined on probabilities $[0, 1]$ as:
    \begin{align}
        \nso \forall x \in D\ (F)\probabilistic &= \prod_{a \in I(D)}\nso F\probabilistic_{\interpretation, \eta[x \rightarrow a]} \\
        \nso F_1 \wedge F_2\probabilistic &= \nso F_1\probabilistic \cdot \nso F_2\probabilistic \\
        \label{eq:probabilistic-expectation}
        \nso x := f(T_1, ..., T_n)\ (F) \probabilistic &= \mathbb{E}_{a\sim \prob_{f}(\cdot|T_1, ..., T_n)} \left[ \nso F\probabilistic_{\interpretation, \eta[x \mapsto a]} \right]
\end{align} 
\end{definition}

In probabilistic semantics, a function $f(x)$ is interpreted as a conditional distribution conditioned on $x$. In this case, we require computing the \emph{expectation} of the formulas being true under the interpreted functions. This happens in Equation~\ref{eq:probabilistic-expectation}. The computation of the expectation depends on whether the output domain $\Omega_{n+1}$ is discrete or continuous. For discrete domains, Equation~\ref{eq:probabilistic-expectation} equals
\begin{equation}
    \label{eq:probabilistic-discrete}
    \nso x := f(T_1, ..., T_n)\ (F) \probabilistic = \displaystyle\sum_{a \in \domainspace_{n+1}}\prob_{f} (a|T_1, ..., T_n) \cdot \nso F\probabilistic_{\interpretation, \eta[x \mapsto a]},
\end{equation}
while for continuous domains it equals
\begin{equation}
    \label{eq:probabilistic-continuous}
    \nso x := f(T_1, ..., T_n)\ (F) \probabilistic = \displaystyle\int_{a \in \domainspace_{n+1}}\prob_{f} (a|T_1, ..., T_n) \cdot \nso F\probabilistic_{\interpretation, \eta[x \mapsto a]} \text{d}a.
\end{equation}
We should note that probabilistic semantics in most practical cases will be intractable because of the exponential recursion introduced in Equation~\ref{eq:probabilistic-discrete}, not to mention the usually intractable integral in Equation~\ref{eq:probabilistic-continuous} \cite{belle2015probabilistic}. We can speed this up with techniques that compile formulas into representations where computing the probability of the formula is tractable \cite{chaviraProbabilisticInferenceWeighted2008,darwicheSDDNewCanonical2011}.
The probabilistic semantics is classical in the limit (Appendix \ref{appendix:classical-in-limit-probabilistic}), and is connected to the standard weighted model counting semantics used in, for example, Semantic Loss \cite{xu2018semanticloss}, SPL \cite{ahmed2022spl} and DeepProbLog \cite{manhaeve2018deepproblog}. See Appendix~\ref{appendix:probabilistic-semantics-semantic-loss} for details.

We can generalise the probabilistic semantics to algebraic model counting \cite{kimmig2017algebraic,derkinderen2024semirings} by considering semirings $\mathcal{B}$ together with a product and a sum operation. This, for example, allows us to compute the most likely assignment to the variables in a formula, or to compute the gradient of the probabilistic semantics using dual numbers.

\subsection{Fuzzy Semantics}
\label{sec:fuzzy-semantics}
The setup for fuzzy semantics is very similar to that of the probabilistic semantics. The two differences are using t-norms and t-conorms to connect fuzzy truth values, and the interpretation of sampling from boolean distributions. 
\begin{definition}
    The fuzzy structure $(\interpretation_F, \eta, [0,1],  \nso\fuzzy)$, where $\interpretation_F$ is an interpretation $I$ except that the predicate symbol $\eval$ is interpreted as the identity function on $[0, 1]$, is defined on fuzzy truth values $[0, 1]$ as:
    \begin{align}
        \label{eq:fuzzy-universal}
        \nso \forall x \in D\ (F)\fuzzy_{\interpretation_F, \eta} &= \bigotimes_{a \in I(D)}\nso F \fuzzy_{\interpretation_F, \eta[x \rightarrow a]} \\
        \nso \exists x \in D\ (F)\fuzzy_{\interpretation_F, \eta} &= \bigoplus_{a \in I(D)}\nso F \fuzzy_{\interpretation_F, \eta[x \rightarrow a]} \\
        \label{eq:fuzzy-and}
        \nso F_1 \wedge F_2\fuzzy& = \nso F_1\fuzzy \tnorm\ \nso F_2\fuzzy, \quad   \nso F_1 \lor F_2\fuzzy = \nso F_1\fuzzy \tconorm\ \nso F_2\fuzzy \\
        \label{eq:fuzzy-eval}
            \nso \eval(x)\fuzzy_{\interpretation, \eta} &= \eta(x), \quad\text{if } \eta(x)\in [0, 1] \\
        \label{eq:fuzzy-statement}
        \nso x \coloneqq f(T_1, ..., T_n) (F) \fuzzy &= \begin{cases} 
            \nso F\fuzzy_{\interpretation_F, \eta[x \mapsto \prob_{f}(1|T_1, ..., T_n)]} &\text{if }\domainspace_{n+1} = \{0, 1\}\\
            \displaystyle\bigoplus_{a \in \domainspace_{n+1}} 
                \prob_{f}(a|T_1, ..., T_n) 
                \otimes \nso F\fuzzy_{\interpretation_F, \eta[x \mapsto a]}
          &\text{if $\Omega_{n+1}$ is finite}
        \end{cases}
    \end{align}
    
    where $\tnorm: [0,1] \times [0,1] \mapsto [0,1]$ is a fuzzy t-norm and $\tconorm: [0,1] \times [0,1] \mapsto [0,1]$ is a fuzzy t-conorm \cite{badreddineLogicTensorNetworks2022,vankriekenAnalyzingDifferentiableFuzzy2022}.

\end{definition}

In the first case of Equation \ref{eq:fuzzy-statement}, fuzzy semantics manipulates distributions over boolean codomains $\Omega_{n+1} = \{0,1\}$ as a single truth value $\prob_{f}(1|T_1, ..., T_n)$.  
The second case is defined for discrete, non boolean codomains.
Fuzzy semantics reasons disjointly over all possible outcomes $a \in \Omega_{n+1}$ by interpreting the probability $\prob_f(a | T_1, \dots, T_n) \in [0,1]$ as truth degrees.
This truth degree is then conjoined with the interpretation of the rest of the formula $F$.
Intuitively, they ask if there \say{exists $a$ such that $f(T_1,\dots,T_n)$ maps to $a$ and that $a$ verifies the rest of the formula $F$}.
We do not give a semantics for continuous or infinite domains in the fuzzy semantics, as we do not know of a standard definition in the neuro-symbolic literature.
The fuzzy semantics is classical in the limit (see Appendix~\ref{appendix:classical-in-limit-fuzzy}), and is closely related to differentiable fuzzy logics such as Logic Tensor Networks \cite{vankriekenAnalyzingDifferentiableFuzzy2022,badreddineLogicTensorNetworks2022} (see Appendix~\ref{appendix:fuzzy-semantics-differentiable-fuzzy}).

In addition to Fuzzy Logics with t-norms and t-conorms for conjunction and disjunction, other NeSy frameworks such as DL2 \cite{fischer2019dl2} and STL \cite{varnai2020robustness} can also be implemented with this semantics. While fuzzy logic acts on truth values in $[0, 1]$, DL2 acts on truth values in $[-\infty, 0]$ and STL in $[-\infty, \infty]$. They choose appropriate differentiable operators to implement the conjunction and disjunction. We refer the reader to \cite{slusarz2023logic} for details.

\subsection{Sampling semantics}
\label{s:semantics_sampling}
The sampling semantics $\nesysystem^ S$ is a simple modification to the classical semantics. It samples a value from each conditional distribution and uses that value to evaluate the formula. Therefore, the only difference in $\nesysystem^ \text{\normalfont S}$ with classical semantics in Definition~\ref{def:classical-semantics} is in Equation \ref{eq:classic-statement}:
\begin{equation}
    \label{eq:sampling-function}
    \nso x := f(T_1, ..., T_n)\ (F) \nscsampling = \nso F\nscsampling_{\interpretation, \eta[x \mapsto \mathsf{sample}(\prob_f(\cdot|T_1, ..., T_n))]} 
\end{equation}
Here, $\mathsf{sample}$ is a (random) function that takes a probability distribution and samples a value from the codomain $\domainspace_{n+1}$ under the distribution $\prob_f(\cdot|T_1, ..., T_n)$. We can repeat the computation of the sampling semantics $\nesysystem^ \text{\normalfont S}$ to reduce variance, like in standard Monte Carlo methods. This semantics can be combined with gradient estimation methods to learn the parameters of neural networks \cite{schulmanGradientEstimationUsing2015,vankriekenStorchasticFrameworkGeneral2021}. A recent implementation of gradient estimation in the context of NeSy is the CatLog derivative trick \cite{de2024differentiable}, but any type of estimator based on the score function (commonly known as REINFORCE) can be used \cite{koolBuyREINFORCESamples2019}. See Appendix~\ref{appendix:gradient-estimation} for a short discussion.

\section{Learning and Reasoning}
\label{sec:learning-with-uller}
This section describes how to use \langname{} for neuro-symbolic learning and reasoning. 
For a learning setting, we extend the definition of an interpretation (Definition \ref{def:interpretation}) to a \emph{parameterised interpretation}. A parameterised implementation allows us to implement neural networks with learnable parameters. For instance, a function $\text{model}()$ can be interpreted as a neural network $I_\btheta(\text{model})= \mathit{NN}_\btheta$. 

\begin{definition}
    A \emph{parameterised interpretation} is an interpretation $I_\btheta$ that is uniquely defined by a set of parameters $\btheta \in \mathbb{R}^ d$.
\end{definition}

Let $F\in \nesylanguage$ denote a \langname{} formula that has a quantifier ranging over a dataset symbol $T$ (for instance Example \ref{ex:composition}).
Learning a parameterised interpretation typically involves searching for an optimal set of parameters $\btheta^* \in \mathbb{R}^ d$ maximising the neuro-symbolic system on $F$ over a dataset $\Omega_T$. In most machine learning settings, we are interested in minimising a loss function over a random minibatch $x_1, ..., x_n\sim \Omega_T$. We can define such a loss function and corresponding minimisation problem with 
\begin{equation}
    L(\btheta) = -\nso F \nsc_{I_{\btheta}\cup \{T\mapsto \{x_1, ..., x_n\}\}, \{\}}, \quad \btheta^ *=\arg\max_{\btheta \in \mathbb{R}^ d} \mathcal{L}(\btheta).
\end{equation}
To allow for minibatching, we interpret the domain symbol $T$ as the minibatch $\{x_1, ..., x_n\}$. We can easily implement variations of this loss. For instance, we can combine multiple formulas and give each different weights. Notice that, for probabilistic and fuzzy semantics, $L(\btheta)$ is differentiable, allowing us to use common optimisers. 
However, not all NeSy structures can be optimised over: This loss only makes sense when a semantics returns a value in an ordered set $\mathcal{B}$, but we also allow NeSy structures to return other kinds of values.  

A different pattern, more related to reasoning, is to find the input $x$ that maximises or minimises the neuro-symbolic system:
\begin{equation}
    x^ * = \arg\max_{x'\in X} \nso F \nsc_{I_\btheta \cup \{T\mapsto \{x\} \}, \{\}}
\end{equation}
This strategy can be combined with adversarial learning to first find the input that most violates the background knowledge, and then corrects that input \cite{minervini-riedel-2018-adversarially}.

\section{Related Work}
\label{sec:related-work}
The last decade has seen the rise of neuro-symbolic frameworks that allow for specifying knowledge about the behaviour of neural networks symbolically \cite{marraStatisticalRelationalNeural2021}. However, unlike \langname{} they are restricted to a single semantics, usually variations of probabilistic (Section~\ref{sec:probabilistic-semantics}) or fuzzy semantics (Section~\ref{sec:fuzzy-semantics}). The majority of current frameworks use the syntactic \emph{neural predicate} construct as discussed in Section~\ref{sec:interpretation}. DeepProbLog \cite{manhaeve2018deepproblog} is a probabilistic logic programming language \cite{deraedtProbabilisticLogicProgramming2015} with neural predicates. Variations of its syntax are used in multiple follow-up works \cite{desmetNeuralProbabilisticLogic2023,wintersDeepstochlogNeuralStochastic2022,maeneSoftUnificationDeepProbabilistic2023}. Scallop \cite{huangScallopProbabilisticDeductive2021} chooses to restrict its language to Datalog to improve scalability, among others \cite{magniniDesignPSyKIPlatform2022}.  For \langname{}, we choose to use an expressive first-order language, leaving scalable inference to the implementation of the NeSy system. Other NeSy frameworks are based on Answer Set Programming \cite{yangNeurASPEmbracingNeural2020,skryaginSLASHEmbracingProbabilistic2021,aspisEmbed2SymScalableNeuroSymbolic2022}, relational languages \cite{pryor2023neupsl,marra2021neural,cohenTensorLogDifferentiableDeductive2016}, temporal logics \cite{umili2023grounding} and description logics \cite{wuDifferentiableFuzzyMathcalALC2022,tang2022falcon,tangTARNeuralLogical2022}, while Logic Tensor Networks \cite{badreddineLogicTensorNetworks2022} is also based on first-order logic, among others \cite{marra2019lyrics,diligentiSemanticbasedRegularizationLearning2017}. Finally, many commonly used NeSy frameworks are restricted to propositional logic \cite{xu2018semanticloss,ahmed2022spl,krieken2023a-nesi,daniele2023refining,GIUNCHIGLIA2024109124,fischer2019dl2}. 

Logic of Differentiable Logics (LDL) \cite{slusarz2023logic} defines a first-order language to compare formal properties of several NeSy frameworks. Compared to \langname{}, LDL is strongly typed, while \langname{} is weakly typed, and LDL does not model probabilistic semantics. In LDL, uncertainty comes from predicates, rather than functions, and does not have a syntactic construct like \langname{}s statement blocks. Pylon \cite{AhmedLTGCKSB022} is a Python library similar in goal to \langname{}. It also allows for expressing propositional logic (CNF) formulas, which can then get compiled into a Semantic Loss or fuzzy loss functions. However, by being restricted to a propositional language, Pylon is limited in expressiveness, and requires the user to manually ground out formulas.

\langname{} is also heavily inspired by probabilistic programming languages \cite{gordon2014probabilistic} such as Stan \cite{carpenter2017stan} that specify probabilistic models in a high-level language. In particular, \langname{} can be considered a first-order probabilistic programming language (FOPPL) \cite{vandemeentIntroductionProbabilisticProgramming2021} defined on boolean outputs. These boolean outputs represent the conditioning (observations) of the probabilistic model. By being first-order, the language is restricted to having a finite number of random variables. Other FOPPL languages centred on neural networks include Pyro \cite{binghamPyroDeepUniversal2019} and ProbTorch \cite{siddharth2017learning}. These languages enforce a probabilistic semantics corresponding to that of \langname{} defined in Section~\ref{sec:probabilistic-semantics}. However, \langname{} does not enforce this semantics and also supports, for instance, fuzzy semantics. We leave an in-depth analysis of the relations between \langname{} and aforementioned probabilistic programming languages for future work. 

Other related work attempts to define building blocks for neuro-symbolic AI \cite{vanharmelenBoxologyDesignPatterns2019a} or to categorise existing approaches \cite{sarkerNeurosymbolicArtificialIntelligence2021a}. We instead focus on a particular set of informed machine learning approaches, and develop a unifying language to allow communicating with them.

\section{Conclusion}
We introduced \langname{}, a \method{}. \langname{} is a first-order logic language designed for neuro-symbolic learning and reasoning, with a special statement syntax for constraining neural networks. We showed how to implement the common fuzzy and probabilistic semantics in \langname{}, allowing for easy comparison between different NeSy systems. 
For future work, we want to implement \langname{} as an easy-to-use Python library to increase the \say{frictionless reproducibility} of NeSy research. In this library, a researcher can easily write and share knowledge, and develop new NeSy benchmarks. We also believe such a library is a good avenue for reducing the barrier of entry into NeSy research.

\section*{Acknowledgements}
We would like to thank 
Frank van Harmelen, Tarek Richard Besold, Luciano Serafini, Antonio Vergari, Pasquale Minervini, Thiviyan Thanapalasingam, Guy van den Broeck, Connor Pryor, Patrick Koopmann, and Mihaela Stoian for fruitful discussions during the writing of this paper. 
This work was supported by the EU H2020 ICT48 project “TAILOR” under contract \#952215. Emile van Krieken was funded by ELIAI (The Edinburgh Laboratory for Integrated Artificial Intelligence), EPSRC (grant no. EP/W002876/1).
\bibliographystyle{splncs04}

\bibliography{bibliography}

\appendix

\section{Practical Examples}
\label{sec:examples}
\EK{This could be an interesting example of statements-in-statements, the 'throw an arbitrary number of dices and check if any are 6s'. 
$$
\forall x\in \mathcal{D}\ \exists y \in \mathbb{Z}: \mathrm{dice}(y)=6\wedge y < \mathrm{count}(x)
$$
\begin{align*}
\forall x\in \mathcal{D}\ ( \\
&c:= \mathrm{count}(x) (\\
&\quad \exists y\in \mathbb{Z} (y < c \wedge ( \\
&\quad \quad d:= \mathrm{dice}(y)\ (d=6)\\
&))))
\end{align*}}

\begin{example}[MNIST Addition]
Suppose we want to express the standard (single-digit) MNIST addition program using \langname{}. In this setting, we have a domain $T$ that represents a training dataset $I(T)$. In Section \ref{sec:learning-with-uller}, we discuss how this training dataset can also be a minibatch of examples. 

Each data point $x$ consists of a pair of images (which we access with the properties $\mathrm{im1}$ and $\mathrm{im2}$) associated to a label representing the value of their sum (which we can intuitively access via the property $\mathrm{sum}$). Finally, we have a function $f$ that we interpret as a neural network classifying MNIST images. Then, if we want to write that for every input the outputs of the neural network should be equal to the sum of the inputs, we can write:
\begin{align*}
    \forall x \in  T & \\
    ( n1 &:= f(x.\mathrm{im1}), n2 := f(x.\mathrm{im2}) \\
       (n1 & + n2 = x.\mathrm{sum}))
\end{align*}
\end{example}

\begin{example}[Smokes Friends Cancer]
    \label{ex:SFC}
    
\def\People{\mathrm{People}}
\def\Friends{\mathrm{Friends}}
\def\Smokes{\mathrm{Smokes}}
\def\Cancer{\mathrm{Cancer}}
\def\Ppl{\mathrm{People}}
\def\Fr{\mathrm{Friends}}
\def\Sm{\mathrm{Smokes}}
\def\Ca{\mathrm{Cancer}}

In this classical example of Statistical Relational Learning introduced by \cite{Richardson2006}, uncertain facts in a population group are modeled using the neural predicates $\Fr(x,y)$ for friendship, $\Sm(x)$ for smoking, and $\Ca(x)$ for cancer. 
As \langname{} relies on functions rather than predicates to model uncertainty, we must use $\eval(a)$ to formalise the problem in our language as explained in Section~\ref{sec:interpretation}. 
For simplicity, we use $(A \lequiv B) \equiv ((A \limplies B) \land (B \limplies A))$ to denote logical equivalences.

Here is an example of a knowledge base for this problem.
Friends of friends are friends:
\begin{align*}
    \forall x &\in \Ppl, y \in \Ppl, z \in  \Ppl \\
                &(\atom_1 := \Fr(x,y), \atom_2 := \Fr(y,z), \atom_3 := \Fr(x,z) \\ 
                &(
                    (\eval(\atom_1) \land \eval(\atom_2)) \limplies \eval(\atom_3)))
\end{align*}

If two people are friends, either both smoke or neither does:
\begin{align*}
    \forall x &\in \Ppl, y \in \Ppl\  \\
    & (\atom_1 := \Fr(x,y), \atom_2 := \Sm(x), \atom_3 := \Sm(y) \\
    & (\eval(\atom_1) \limplies (\eval(\atom_2) \lequiv \eval(\atom_3)) ) )
\end{align*}

Friendless people smoke:
\begin{align*}
    \forall x & \in \Ppl \\
    & (\neg \exists y \in \Ppl\ (\atom_1 := \Fr(x,y)(\eval(\atom_1))) \\
    & \phantom{(} \limplies \atom_2 := \Sm(x)(\eval(\atom_2))
    )
\end{align*}

Smoking causes cancer:
\begin{align*}
    \forall x & \in \Ppl \\
    & ( \atom_1 := \Sm(x), \atom_2 :=  \Ca(x) \\
    & ( \eval(\atom_1) \limplies \eval(\atom_2) )
)
\end{align*}

Notice that, according to the definitions of Section \ref{sec:interpretation}, the probabilistic interpretation of the above formula will assume conditional independence between $a_1 \sim \prob_\Sm(\cdot | x)$ and $a_2 \sim \prob_\Ca(\cdot | x)$.
To model a dependence of cancer on smoking, i.e. $a_2 \sim \prob_\Ca(\cdot | x, a_1)$, we can make the probability explicitly depend on the previous variable:
\begin{align*}
    \forall x & \in \Ppl \\
        & ( \atom_1 := \Sm(x), \atom_2 :=  \Ca(x, \atom_1) \\
        & ( \eval(\atom_1) \limplies \eval(\atom_2) )
    )
\end{align*}

Next, we have labelled examples for each relationship. For example, for $\Friends()$, drawing examples from a dataset $T_{\Fr}$:
\begin{align*}
    \forall t & \in T_{\Fr} \\
        & ( l := \Fr(t.\mathrm{x}, t.\mathrm{y}) \\
        & ( l = t.\mathrm{label} )
    )
\end{align*}
\end{example}

\section{Gradient estimation}
\label{appendix:gradient-estimation}
The sampling semantics in Equation~\ref{eq:sampling-function} is a simple way to estimate the truth value of a formula. However, since sampling is not a differentiable operation, it is not possible to use this semantics to train the neural networks. Instead, we can use the score function gradient estimation method \cite{schulmanGradientEstimationUsing2015} to estimate the gradient of the truth value of a formula with respect to the parameters of the neural networks. However, this requires adapting the evaluation of the formula to incorporate score function terms. We give a brief description of how one might go about doing that in Appendix \ref{appendix:gradient-estimation}.

One way to implement gradient estimation methods for simple \langname{} programs is to use the DiCE estimator \cite{foersterDiCEInfinitelyDifferentiable2018} which introduces the \emph{MagicBox} operator $\magic(x)=\exp(x-\bot(x))$, where $\bot$ is the \textsf{StopGradient} operator used in deep learning frameworks. This operator allows us to add a term that only appears when we differentiate it, and equals 1 during the forward pass. To incorporate DiCE for \method{}, we have to modify Equation \ref{eq:classic-statement}
\begin{equation}
    \label{eq:dice-statement}
    \nso x:= f(T_1,\dots,T_n)(F)\ \nscsampling = \nso F\classic_{\interpretation, \assign(\eta, S)} \cdot \sum_{i=1}^ n \magic(\log \prob_{f_i}(\assign(\eta, S)[x_i]))
\end{equation}

Extensions of the DiCE estimator can be used to implement a wide variety of gradient estimation methods \cite{vankriekenStorchasticFrameworkGeneral2021}. 

\section{Classical in the limit}
\subsection{Probabilistic semantics}
\label{appendix:classical-in-limit-probabilistic}
The probabilistic semantics is \emph{classical in the limit}. To show this, we note that we require that the domain becomes $\{0, 1\}$ instead of probabilities $[0, 1]$. Under this domain, the product is equal to the $\min$ function. We can use this to rewrite all but the interpretation of statements into the classical semantics.

Next, take for a statement $x := f(T_1, ..., T_n)(F)$ the induction assumption that $\nso F \probabilistic_{\hat{I}, \eta} = \nso F \classic_{I, \eta}$, where $\hat{I}$ is defined as in Definition~\ref{def:classical-in-limit}. Then the interpretation of a statement is: 
\begin{align*}
   \mathbb{E}_{a\sim \prob_{\hat{f}}(\cdot|T_1, ..., T_n)}[\nso F\probabilistic_{\hat{\interpretation}, \eta[x \mapsto a]}] 
    =\nso F\probabilistic_{\hat{\interpretation}, \eta[x \mapsto \arg\max_{a\in \Omega_{n+1}}\prob_f(a|T_1, ..., T_n))]}
\end{align*}
Here, we reduce the expectation by noting that since $p_{\hat{f}}(a|T_1, ..., T_n) =  \delta(a-\arg\max_{a'} \prob_f(a'|T_1, ..., T_n))$, exactly one element gets 1 probability (or a single element with non-zero probability, in the case of continuous distributions). This single element is chosen on the right side. 
Then, we use the induction assumption to find that this is equal to the classical semantics of statements given in Equation~\ref{eq:classic-statement}. 

\subsection{Fuzzy semantics}
\label{appendix:classical-in-limit-fuzzy}
Using the axioms of t-norms, we find that the fuzzy semantics is also classical in the limit. This again can be proven by induction. For Equations~\ref{eq:fuzzy-universal} and \ref{eq:fuzzy-and}, we use the boundary conditions of t-norms, which states that $x \tnorm 1=x$ for $x\in [0, 1]$. Therefore, if $x=0$, $0 \tnorm 1=0$ and if $x=1$, $1 \tnorm 1=1$, meaning t-norms act as the $\min$ operator under the domain $\{0, 1\}$. 

Next, consider the program fragment $x := f(T_1, ..., T_n)\ (F)$ and take the induction assumption $\nso F \fuzzy_{\hat{I}, \eta} = \nso F \classic_{I, \eta}$. First, assume the domain $\Omega_{n+1} = \{0, 1\}$ and assume $\arg\max_{a\in \{0, 1\}} \prob_f(a|T_1, ..., T_n)=1$. Then, the interpretation of the statement is $\nso F\fuzzy_{\hat{\interpretation}, \eta[x \mapsto \prob_{\hat{f}}(a=1|T_1, ..., T_n)]}=\nso F\fuzzy_{\hat{\interpretation}, \eta[x \mapsto 1]}$, since the Dirac distribution will put all its mass on the output $1$. Similarly, if $\arg\max_{a\in \{0, 1\}} \prob_f(a|T_1, ..., T_n)=0$, then the interpretation is $\nso F\fuzzy_{\hat{\interpretation}, \eta[x \mapsto 0]}$. Then we can simply use the induction assumption.

Finally, if $\Omega_{n+1} \neq \{0, 1\}$, then we know there is a unique output $a\in \Omega_{n+1}$ such that $p_{\hat{f}}(a|T_1, ..., T_n)=1$, while for the other outputs $p_{\hat{f}}(a|T_1, ..., T_n)=0$. Then, using associativity and commutativity of the t-conorm $\oplus$, the interpretation of the statement is 
\begin{align*}
    &\nso F\fuzzy_{\hat{\interpretation}, \eta[x \mapsto a]}\tnorm \prob_{\hat{f}}(a|T_1, ..., T_n) \oplus \bigoplus_{a'\in \Omega_{n+1}\setminus \{a\}} \nso F\fuzzy_{\hat{\interpretation}, \eta[x \mapsto a']}\tnorm \prob_{\hat{f}}(a'|T_1, ..., T_n) \\
    &\nso F\fuzzy_{\hat{\interpretation}, \eta[x \mapsto a]}\tnorm 1 \oplus \bigoplus_{a'\in \Omega_{n+1}\setminus \{a\}} \nso F\fuzzy_{\hat{\interpretation}, \eta[x \mapsto a']}\tnorm 0 \\
    &\nso F\fuzzy_{\hat{\interpretation}, \eta[x \mapsto a]}\oplus \bigoplus_{a'\in \Omega_{n+1}\setminus \{a\}}  0 = \nso F\fuzzy_{\hat{\interpretation}, \eta[x \mapsto a]}
\end{align*}
where we again use the boundary conditions of the t-norm $\tnorm$ ($1\tnorm x = x$) and t-conorm ($0\oplus x = x$). 

\section{Relation of Probabilistic semantics to the Semantic Loss}
\label{appendix:probabilistic-semantics-semantic-loss}
Here, we show why the probabilistic semantics is equivalent to the weighted model counting semantics used in, for instance, the Semantic Loss. Let $F$ be a closed formula without any statements $x:=f(T_1,\dots,T_n)(F')$ that only involves variables $x_1, ..., x_n$ over finite domains. The \emph{weighted model count (WMC)} is the evaluation of the classical semantics weighted by probabilities of the assignments to variables. These probabilities are often assumed to be independent, although our framework also allows for the probabilities to depend on previous variables. This is illustrated in Example \ref{ex:SFC}. The definition of the WMC is
\begin{equation}
    \begin{aligned}
    \mathsf{WMC} &= \sum_{a_1 \in \domainspace_1} ... \sum_{a_n \in \domainspace_n} \prod_{i=1}^n \prob_{f_i}(a_i) \nso F\classic_{\interpretation, \{x_1 \mapsto a_1, ..., x_n \mapsto a_n\}} \\
    &= \sum_{a_1 \in \domainspace_1} \prob_{f_1}(a_1) ... \sum_{a_n \in \domainspace_n} \prob_{f_n}(a_n) \nso F\classic_{\interpretation, \{x_1 \mapsto a_1, ..., x_n \mapsto a_n\}} .
    \end{aligned}
\end{equation}
Next, we rewrite this into a program $x_1 := f_1(), ..., x_n := f_n()\ (F)$ such that the probabilistic semantics in Definition \ref{def:probabilistic-semantics} is equal to the weighted model count. 
For ease of notation, let us denote $S_i$ each statement $x_i := f_i()$ for $i=1, ..., n$. Then, we find the probabilistic semantics of the program by sequentially expanding the interpretation of the statements:
\begin{equation}
    \begin{aligned}
        \nso S_1, ..., S_n (F)\probabilistic_{I, \{\}} &=  \sum_{a_1\in \domainspace_1} \prob_{f_1}(a_1) \cdot \nso S_2, ..., S_n (F) \probabilistic_{I, \{ x_1 \mapsto a_1\}} \\
        &\dots \\
        &= \sum_{a_1 \in \domainspace_1} \prob_{f_1}(a_1) ... \sum_{a_n \in \domainspace_n} \prob_{f_n}(a_n) \nso F\classic_{\interpretation, \{x_1 \mapsto a_1, ..., x_n \mapsto a_n\}} \\
        &= \mathsf{WMC}
    \end{aligned}
\end{equation}
where in the last step we use that since the domains are finite and $F$ does not contain statements, the probabilistic semantics of $F$ is equal to the classic one. 

\section{Relation of Fuzzy Semantics to Differentiable Fuzzy Logics}
\label{appendix:fuzzy-semantics-differentiable-fuzzy}
Fuzzy logics are actively used in NeSy \cite{badreddineLogicTensorNetworks2022,vankriekenAnalyzingDifferentiableFuzzy2022,daniele2023refining,GIUNCHIGLIA2024109124}. We show how existing NeSy systems using fuzzy logics arise from the fuzzy semantics of \langname{}.
Existing fuzzy logics systems align with our interpretations of terms and logical operators, 
but differ in their use of fuzzy predicates, which are interpreted as functions to $[0, 1]$, that is, $I_\mathrm{NeSy}(P): \domainspace_1 \times \dots \times \domainspace_n \rightarrow [0, 1]$. Then, the truth value of a formula is computed by evaluating the formula with the fuzzy semantics.

We can emulate this in our fuzzy semantics with the $\eval()$ predicate and proof by induction. For each neural predicate $I_\mathrm{NeSy}(P_i): \domainspace^i_{1} \times \dots \times \domainspace^i_{n_i} \rightarrow [0,1]$, we define a \langname{} function $I(f_i): \domainspace^i_{1} \times \dots \times \domainspace^i_{n_i} \rightarrow (\{0,1\} \rightarrow [0,1])$ such that:
\begin{equation}
    \label{eq:fun_to_neural_pred}
    I_\mathrm{NeSy}(P_i)(T^i_1, \dots, T^i_{n_i}) = I(f_i)(T^i_1, \dots, T^i_{n_i})(1)
\end{equation}
Let $F$ be a first-order logic formula with no statements nor functions, and $\nso F \nesyfuzzy$ be its interpretation in a fuzzy NeSy system. 
Let $F$ contain $k$ neural atoms $P_i(T^i_1, \dots, T^i_{n_i})$, $i = 1 \dots k$.
Let $S_1, \dots, S_k\ (F')$ be a \langname{} program with $k$ statements where $S_i$ defines $x_i := f_i(T^i_1, \dots, T^i_{n_i})$, $i=1,\dots,k$, and $F'$ is $F$ where we replace every mention of $P_i(T^i_1, \dots, T^i_{n_i})$ by $\eval(x_i)$.
We have:
\begin{align}
    \nso S_1, \dots, S_k\ (F')\fuzzy 
        &= \nso F' \fuzzy_{I, \eta[x_1 \mapsto p_{f_1}(1 | T^1_1, \dots, T^1_{n_1}), \dots, x_k \mapsto p_{f_k}(1 | T^k_1, \dots, T^k_{n_k})]}\\
        &= \nso F' \fuzzy_{I, \eta[x_1 \mapsto I_\mathrm{NeSy}(P_1)(T^1_1, \dots, T^1_{n_1}), \dots, x_k \mapsto I_\mathrm{NeSy}(P_k)(T^k_1, \dots, T^k_{n_k})]} \label{eq:using_fun_to_neural_pred}\\
        &= \nso F \nesyfuzzy \label{eq:using_F'}
\end{align}
Equality \eqref{eq:using_fun_to_neural_pred} stems from definition \eqref{eq:fun_to_neural_pred}.
We derive equality \eqref{eq:using_F'} by induction.
First, note that according to the definition of $I(\eval)$ and the assignment in \eqref{eq:using_fun_to_neural_pred}, we have:
\begin{equation}
    \label{eq:fuzzy_neuralpred_interpreted_like_eval}
    \nso \eval(x_i) \fuzzy = I_\mathrm{NeSy}(P_i)(T^i_1, \dots, T^i_{n_i}) = \nso P(T^i_1, \dots, T^i_{n_i}) \nesyfuzzy \quad \text{for } i=1,\dots,k
\end{equation}
If our semantics use the same t-norm operator $\tnorm$ as the NeSy system, then: 
\begin{equation}
    \nso F_1 \land F_2 \fuzzy = \nso F_1 \fuzzy \tnorm \nso F_2 \fuzzy 
    = \nso F_1' \nesyfuzzy \tnorm \nso F_2' \nesyfuzzy
    = \nso F_1' \land F_2' \nesyfuzzy 
\end{equation}
where in the second equality we use the induction hypothesis $ \nso F_1 \fuzzy =  \nso F_1' \nesyfuzzy $ and  $\nso F_2 \fuzzy =  \nso F_2' \nesyfuzzy $.
The same can naturally be derived for other logical connectives.
It follows that we can emulate any formula $F$ built with the neural predicates $P(T^i_1, \dots, T^i_{n_i})$, by building formula $F'$ with the equivalently interpreted $\eval(x_i)$ (see Equation \eqref{eq:using_fun_to_neural_pred}) and the same logical constructs, such that $\nso F \nesyfuzzy = \nso S_1, \dots, S_k\ (F') \fuzzy$.

\end{document}